\documentclass{article}

\usepackage{arxiv}

\usepackage[utf8]{inputenc} 
\usepackage[T1]{fontenc}    
\usepackage{hyperref}       
\usepackage{url}            
\usepackage{booktabs}       
\usepackage{amsfonts}       
\usepackage{nicefrac}       
\usepackage{microtype}      
\usepackage{lipsum}
\usepackage{graphicx}
\usepackage{float}
\usepackage{natbib}
\setcitestyle{square,sort,comma,numbers}

\title{Non-Intrusive Load Monitoring with Fully Convolutional Networks}

\author{
  Cillian Brewitt\\\
  School of Informatics\\
  University of Edinburgh\\
  United Kingdom \\
  \texttt{cillian.brewitt@ed.ac.uk} \\
   \And
 Nigel Goddard \\
  School of Informatics\\
  University of Edinburgh\\
  United Kingdom \\
  \texttt{nigel.goddard@ed.ac.uk} \\
}

\begin{document}
\maketitle

\begin{abstract}
Non-intrusive load monitoring or energy disaggregation involves estimating the power consumption of individual appliances from measurements of the total power consumption of a home. Deep neural networks have been shown to be effective for energy disaggregation. In this work, we present a deep neural network architecture which achieves state of the art disaggregation performance with substantially improved computational efficiency, reducing model training time by a factor of 32 and prediction time by a factor of 43. This improvement in efficiency could be especially useful for applications where disaggregation must be performed in home on lower power devices, or for research experiments which involve training a large number of models.

\end{abstract}


\section{Introduction}

Non-intrusive load monitoring (NILM) or electricity disaggregation is the task of estimating the power consumption of appliances in a home using only measurements of the power usage for the entire home. One application of NILM is to produce a summary of how total energy use is spread across different appliances using readings from a single smart meter. This could help household occupants to make informed decisions in order to to reduce their energy consumption. Research has shown that giving an appliance level breakdown of energy usage to users can lead to a 5\%-15\% reduction in total energy consumption \cite{fischer08}. This can both save the user money on their electricity bill, and have an environmental benefit by reducing the use of electricity, much of which is produced from fossil fuels. Another application of NILM is demand response in smart grids \cite{demandresponse11}. An alternative way of tackling these problems would be to install a separate electricity measurement device for each appliance, however this would be quite costly, so NILM is generally preferable.

Although NILM was first introduced by G. W. Hart in 1992 \cite{nilm92}, there has recently been a surge of interest in NILM due to the widespread introduction of smart meters, and many new algorithms and techniques have been developed. NILM is a single-channel blind source separation (BSS) problem, where multiple sources must be extracted from a single channel which is a linear combination of the sources. Other examples of single channel BSS problems include audio and speech separation \cite{bss17}. These tasks are similar enough that advances in models used for one form of single channel BSS could be applied to other forms. NILM is a difficult task, with many sources of uncertainty. The readings from electricity meters can be noisy, multiple appliances can change states simultaneously, and different appliances sometimes have similar power consumption. For these reasons, machine learning is quite well suited to the task of NILM. Some existing approaches to NILM include factorial hidden Markov models (FHMM) \cite{zhong14}, latent Bayesian melding (LBM) \cite{lbm15}, and graph based signal processing \cite{zhao15} \cite{zhao16}. These techniques require domain specific features to be extracted from the raw data, such as state changes and durations. Another approach to NILM is to use deep neural networks \cite{Kelly15} \cite{pointnet17} \cite{chen18}. Neural networks can automatically learn hierarchies of features from the raw electricity meter readings, so no domain specific feature engineering is required. Both recurrent neural networks (RNN) and convolutional neural networks (ConvNet) have been applied to this task \cite{Kelly15}.

The main contribution of this work is to apply a new neural network architecture to the task of energy disaggregation which drastically reduces training and prediction time compared to existing approaches, while achieving state of the art performance. If NILM is to be used at scale to disaggregate electricity data from residential properties numbering in the millions, as some applications of the data would imply, then reducing training and prediction times becomes an important goal. We compared our approach to an existing ConvNet for energy disaggregation, the "sequence to point" model used by Zhang et al. \cite{pointnet17} and showed that our model achieves state-of-the-art performance. For the comparison we used several metrics, which measure both how accurate and how computationally efficient the models were.

\section{Electricity Disaggregation}

The task of electricity disaggregation can be described as follows. Let $Y=\{y_1,y_2,...,y_T\}$ represent the total (aggregate) power usage of a household over a time period of length $T$. Suppose that there are $I$ different appliances in the household, where the power usage of appliance $i$ is represented by $X_i=\{x_{i1},x_{i2},...,x_{iT}\}$. The aggregate power $y_t$ at each time step is assumed to be the sum of the power usage of each of the individual appliances, plus some Gaussian noise $\epsilon_t$ with zero mean. For most applications, we are only interested in determining the power usage of a subset of all of the appliances, and the other appliances can be modelled as an unknown factor $u=(u_1,u_2,...,u_T)$. Putting these terms together gives us the full model of the system as shown in equation \ref{nilmeq}:

\begin{equation} \label{nilmeq}
y_t=\sum^{I}_{i=1}x_{it}+u_t+\epsilon_t
\end{equation}

\section{Related Work}

In recent years, deep neural networks (DNN) \cite{deeplearning15} have seen an explosion of usage. This has largely been driven by their success in areas such as computer vision \cite{alexnet12}, natural language processing \cite{nlp11}, and speech recognition \cite{dnnspeech12}. Most other forms of machine learning require domain specific features to be designed to extract information from raw data. However, the main advantage of DNNs is that they automatically learn to extract a hierarchy of features from raw high dimensional data such as pixels in an image, and they do not require any domain specific feature engineering. Convolutional neural networks (ConvNets) are a type of DNN that are designed to process data that has a spatial (or in our case temporal) interpretation, where translation invariance is important. ConvNets were originally used for image classification \cite{lecun1998}, and have since been applied to many other areas, including speech processing \cite{wavenet16} and NILM \cite{Kelly15}. Another form of DNN are recurrent neural networks (RNN) \cite{rnn94}. RNNs are usually used for processing temporal sequences, and in contrast to ConvNets, they contain recurrent connections which allow the network to retain a memory of past events.

Kelly and Knottenbelt \cite{Kelly15} tried several different approaches to using DNNs for NILM, using both ConvNets and RNNs. Sequence to sequence (S2S) models were used in each case, with a sliding time window approach used to disaggregate long sequences. Each model took a fixed size time window of aggregate electricity usage as input and gave a window of inferred appliance power usage as output. Out of all of the models that were compared, the most effective approach was found to be a denoising autoencoder, which was a ConvNet that took a sequence of aggregate power usage readings as an input and then output a sequence of the same length giving predictions of the disaggregated power usage of an appliance. This approach was found to outperform previously used NILM algorithms such as AFHMM. One aspect of S2S models is that when overlapping windows are used, multiple predictions will be made for each element of the output sequence. This is usually handled by averaging all of the predictions for each element. Another drawback of S2S models is that some predictions made for each element may be more accurate than others. For predicted elements that are near the midpoint of the predicted window, the model can make use of all of the nearby elements of the input sequence. However, for predicted elements that are near the edges of the window, the model cannot use all of the nearby elements of the input sequence. 

Another approach to using a DNN for NILM is a sequence to point (S2P) model, where a fixed length sliding time window of aggregate power readings is taken as input, and the output of the model is a prediction of the appliance power usage at a single point in time, typically at the midpoint of the input window. Zhang et al. \cite{pointnet17} showed that a S2P model outperformed the S2S architecture used by Kelly and Knottenbelt, and also found that an S2P model still outperformed and S2S model when similar network architectures were used for both models. One advantage of an S2P model is that only a single prediction is made for each element of the output sequence. This allows all of the representational power of the model to be focused on a predicting a single element, and multiple predictions do not need to be averaged. 

Several other neural network architectures for NILM have also been proposed \cite{pedro16} \cite{chen18}. Chen et al. \cite{chen18} presented a new model architecture, which was found to outperform the model used by Zhang et al. \cite{pointnet17}. The model used by Chen et al. included max pooling layers, convolutional blocks with gated linear units (GLU) and residual blocks, which may have helped to improve the accuracy of the model. Another difference between this model and previous S2S models is that the output window length $l_{out}$ used was much shorter than the input window length $l_{in}$. This may have partially alleviated the problem discussed above where some of the output elements of a S2S model with $l_{out}=l_{in}$ do not have access to all of the nearby input elements. 

\section{Network Architecture}

\subsection{Fully Convolutional Network}

All of the models described above involve using overlapping sliding windows.  When using these approaches, the same segments of the input sequence are processed many times as part of different input windows, which is somewhat redundant. To overcome this redundancy and speed up training and prediction, we propose to use a fully convolutional network (FCN) for disaggregation. FCNs, originally known as Space Displacement Neural Networks, have previously been applied to dense prediction tasks such as image segmentation \cite{Shelhamer17} \cite{matan92}. ConvNets usually consist of a series of convolutional layers followed by one or more fully connected layers. In contrast to this, FCNs do not have any fully connected layers and instead only consist of a series of convolutional layers. The output of the model is the feature map from the final convolutional layer. These models share some similarities with both the S2S and S2P models described above. Similarly to the S2S model, the FCN takes a sequence as input and gives a sequence as output. However, similarly to the S2P model, each element in the output sequence depends only on a window of the input sequence centred around that output element. This is because the output of each neuron in a convolutional layer is dependent only on a local region of the input sequence known as its local receptive field \cite{Luo16}. Since the outputs of the FCN are translation invariant, averaging multiple predictions from overlapping sliding windows would be redundant. 

\subsection{Dilated Convolutions}

Another difference in our model compared to previous ConvNets used for energy disaggregation is that we used dilated convolutions. ConvNets with dilated convolutions have recently been shown to be effective for dense prediction tasks such as image segmentation \cite{dialatedconv15} and speech synthesis \cite{wavenet16}. Dilated convolutions are essentially convolutions with gaps, and the dilation rate is the number of elements from the input sequence which is skipped with each gap. This allows a filter of a certain size to cover a larger section of the input sequence as the dilation rate is increased. Stacking a series of layers with dilated convolutions with an increasing dilation rate allows ConvNets to extract contextual information from multiple time scales without losing resolution. If the dilation rate of sequential convolutional layers is increased exponentially, then the receptive field can be increased exponentially with only a linear increase in network complexity. The filters in the final layer of the model proposed in this work had a receptive field of 2053 samples, equivalent to a 274 minute window of the input sequence.

\begin{figure}[h]
\label{network-architecture}
\caption{Neural network architecture}
\centering\includegraphics[width=0.8\linewidth]{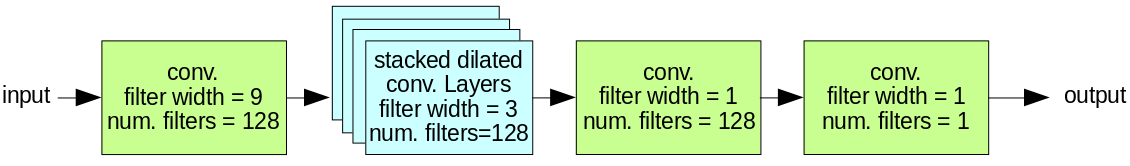}

\end{figure}

\subsection{Fully Convolutional Model}
The neural network architecture used is shown in figure 1. All of the layers except the final layer had 128 filters, which was selected by hand as the highest number of filters which did not substantially slow down training. The initial layer was a convolutional layer with a filter width of 9. A larger filter width was used here than in the following layers in order to better balance the number of parameters in each layer. This was followed by a series of 9 dilated convolutional layers. Each dilated layer had a filter width of 3, as this is the smallest filter than can take past, present and future information into account. The dilation rate of the first dilated convolutional layer was set to 2, and the dilation rate of each subsequent dilated convolutional layer had a dilation rate twice that of the previous layer. This allowed the receptive field of the network to be increased exponentially while linearly increasing the number of layers. Following the layers with dilated convolutions, a convolutional layer with 128 filters and a width of 1 was used to allow the output of the prior layers to be further refined. The final layer had a single filter of width 1 in order to reduce the output of the network to a single channel. 

\section{Experiments}

In this section, we describe how a comparison was carried out between an existing state-of-the-art S2P ConvNet model and our proposed FCN model. We then present the results achieved by each model.

\subsection{Dataset}

The IDEAL dataset includes energy usage data from UK homes. It includes the aggregate apparent power usage of all homes sampled at 1 second intervals. A subset of the homes, which are the ones used in this study, also have appliance level submetering. Some appliances in these homes had their active power usage measured at 1 second intervals, and other appliances had their power usage metered at 5 second intervals.

Readings from a large number of appliance types are included in the dataset. However, many of these appliances only occured in few homes with submetered data, or did not consume a significant amount of energy. The following appliance types were selected for modelling: washing machine, microwave, dishwasher, electric kettle, electric shower, and electric cooker. These appliance types were selected because they all have particularly high energy usage and are commonly found in homes in the IDEAL dataset, as can be seen in table \ref{appliance_counts}.

We used data detailing the energy use of 34 homes from the IDEAL dataset. This included all of the homes which had  at least 30 days of submetered data for at least one of the appliance types mentioned above. The duration of readings for these homes ranged from 39 days to 627 days, using data collected up to and including 30 June 2018. These 34 homes were divided into a training set with 18 homes, a validation set with 8 homes, and a test set with 8 homes, as shown in table \ref{dataset_homes}. The sets were selected by random search so that the ratio between the counts of each appliance type within each set were as similar as possible across all sets. The count of each appliance type in each set is shown in table \ref{appliance_counts}.

\begin{table}[h]
\caption{The home id of homes in the training, validation, and test sets.}
\label{dataset_homes}
\begin{tabular}{|l|l|}
\hline
           & \textbf{Homes}                                                                                                                       \\ \hline
Training   & \begin{tabular}[c]{@{}l@{}} 62, 65, 96, 105, 106, 128, 136, 145, 162, 168,\\ 169, 175, 228, 231, 238, 255, 263, 328\end{tabular} \\ \hline
Validation & 61, 63, 139, 140, 146, 208, 225, 268                                                                                            \\ \hline
Test       & 73, 171, 212, 227, 242, 249, 264, 266                                                                                           \\ \hline
\end{tabular}
\end{table}

\begin{table}[h]
\caption{The count of each appliance type in the training, validation, and test sets.}
\label{appliance_counts}
\begin{tabular}{|l|l|l|l|l|}
\hline
\textbf{Appliance Type} & \textbf{Training} & \textbf{Validation} & \textbf{Test} & \textbf{Total} \\ \hline
Washing Machine & 13           & 3              & 5        & 21    \\
Microwave       & 12           & 4              & 5        & 21    \\
Dishwasher      & 10           & 3              & 4        & 17    \\
Electric Kettle          & 8            & 3              & 4        & 15    \\
Electric Shower & 6            & 2              & 3        & 11    \\
Electric Cooker & 5            & 1              & 2        & 8    \\
\hline
\end{tabular}
\end{table}

\subsection{Data Preprocessing}

We carried out some preprocessing on the raw data to make it suitable for training and evaluating NILM models. The data was initially cleaned and resampled as described below. Following this, the data was windowed and normalised/standardised using slightly different methods for each type of model.

Each home in the IDEAL dataset had two sensors measuring the instantaneous apparent root mean square (RMS) power usage for the whole home at 1Hz, one which was rated for 30A and one which was rate for 100A. The readings from each of these were merged, using the readings from the 100A sensor when the power usage was greater than 4kW, and using readings from the 30A sensor otherwise, reflecting the relative manufactured measurement accuracy of the two sensors in different power ranges. There were some longer gaps of missing data between readings due to signal propagation issues and system downtime. Gaps in readings for the 30A sensor were filled with readings from the 100A where possible. Other gaps in readings shorter than one minute were filled with the previous valid value, and periods of time containing longer gaps were excluded when training and testing models. 

Individual appliance monitors (IAMs) were used to measure the active power usage of certain appliances at sampling intervals of either 1 second or 5 seconds. The IAMs for kettles, microwaves, dishwashers, and washing machines used a sampling interval of 1 second, while the IAMs for electric showers and electric cookers used a sampling interval of 5 seconds. To reduce the amount of data, readings were only reported from the IAMs when the appliance power usage changed, or at least once per hour if the appliance power usage did not change. Because of this, gaps in readings with a duration of less than or equal to one hour were assumed to be periods of time during which the appliance power usage did not change, and were forward filled. Gaps longer than one hour were assumed to be due to missing data and were excluded when training and testing models. Before training and testing, all sensor readings were downsampled to a sampling interval of 8 seconds by taking the mean power usage across each 8 second interval.

Some sensors exhibited some noise in the form of erroneous spikes. Spikes in power above a certain threshold were set to zero, with a different threshold used for each type of sensor. The thresholds used were 20kW for aggregate readings, 15kW for electric shower and electric cooker IAMs, and 4kW for all other IAMs.

\subsubsection{Preprocessing for Sequence to Point Model}

Neural networks tend to learn more efficiently if the input is scaled to have zero mean and unit standard deviation. Before training and evaluating the S2P model, the data was standardised and windowed in similar way to that used by Zhang et al. \cite{pointnet17}. The aggregate data was standardised by subtracting the mean and dividing by the standard deviation of the aggregate power from the entire training set. The targets (appliance power usage) were standardised by subtracting the mean and dividing by the standard deviation of the power used for each appliance type while switched on.

\subsubsection{Preprocessing for Fully Convolutional Network}

Following the approach used by Kelly and Knottenbelt \cite{Kelly15}, we divided each input sequence by the standard deviation of all inputs from the training set, and divided each input sequence by its own mean, rather than the mean from the entire training set. The targets were normalised by dividing by the mean on power for each appliance type.

Although fully convolutional networks (FCN) can theoretically take an input sequence of any length, we divided the inputs and targets into shorter windows to make training practical. An output window length of 2053 and an input window of length of 4105 was used to ensure that the entire receptive field of each filter in the final convolutional layer was included in the input. The input and output sequences were separated into windows such that the output windows did not overlap and the input windows partially overlapped.

\subsection{Model Training}

The S2P models were trained using the same methods as used by Zhang et al. \cite{pointnet17}. These models were trained using the ADAM \cite{KingmaB14} optimization method, with a learning rate of $1\times10^{-3}$, and a batch size of 4096. Training was terminated after the validation error had not improved for 10 epochs, or after 100 total epochs.

The FCN models were also trained using the ADAM optimization method. An initial learning rate of $1\times10^{-3}$ was used, which was reduced by a factor of 10 each time the validation error rate had not improved for 10 epochs. Training was terminated when the validation error had not improved for 15 epochs, or after a total of 200 epochs. A batch size of 256 was used.

\subsection{Performance Evaluation}

We evaluated the performance of the models using two different metrics, both of which are commonly used to evaluate NILM models. One performance metric used was mean absolute error (MAE). This is given by equation \ref{mae}, where $x_t$ is the ground truth appliance power consumption, and $\hat{x}_t$ is the predicted appliance power consumption at time $t$. This metric is useful if we are interested at the error in predicted power at every point in time.

\begin{equation} \label{mae}
MAE= \frac{1}{T}\sum_{t=1}^T |\hat{x}_t-x_t|
\end{equation}

For some applications, the error at every point in time is not important, and we are only interested in the error in total energy use over longer periods of time. The normalised signal aggregate error (SAE) \cite{zhong14} measures the error in total energy consumption over the entire available time period for each house, as shown in equation \ref{sae}. 

\begin{equation} \label{sae}
SAE= \frac{|\sum_t \hat{x}_t - \sum_t x_t|}{\sum_t x_t}
\end{equation}

In order to compare the computational efficiency of each model, we also recorded the time taken for training and prediction, and the number of model parameters.

\subsection{Results}

In most cases the FCN achieved a lower error than the S2P model, as can been seen in table \ref{tab:results}. The FCN achieved a lower MAE than the S2P model for all appliance types, and reduced the mean MAE from 34.8 watts to 14.9 watts. The FCN also reduced the mean SAE from 0.67 to 0.33, however the S2P model still achieved a lower SAE for some appliances. The error for the S2P model for the electric shower is especially high. Upon examination of the predictions made by the S2P model for the shower, it could be seen that this was because the model consistently predicted power consumption of 50 - 100 watts when the shower was not in use. The exact cause of this is difficult to investigate due to the black-box nature of deep neural networks. Some examples of the predictions made by each model are shown in figure 2.

\begin{figure}[h]
\label{fig:examples}
\centering\includegraphics[width=1\linewidth]{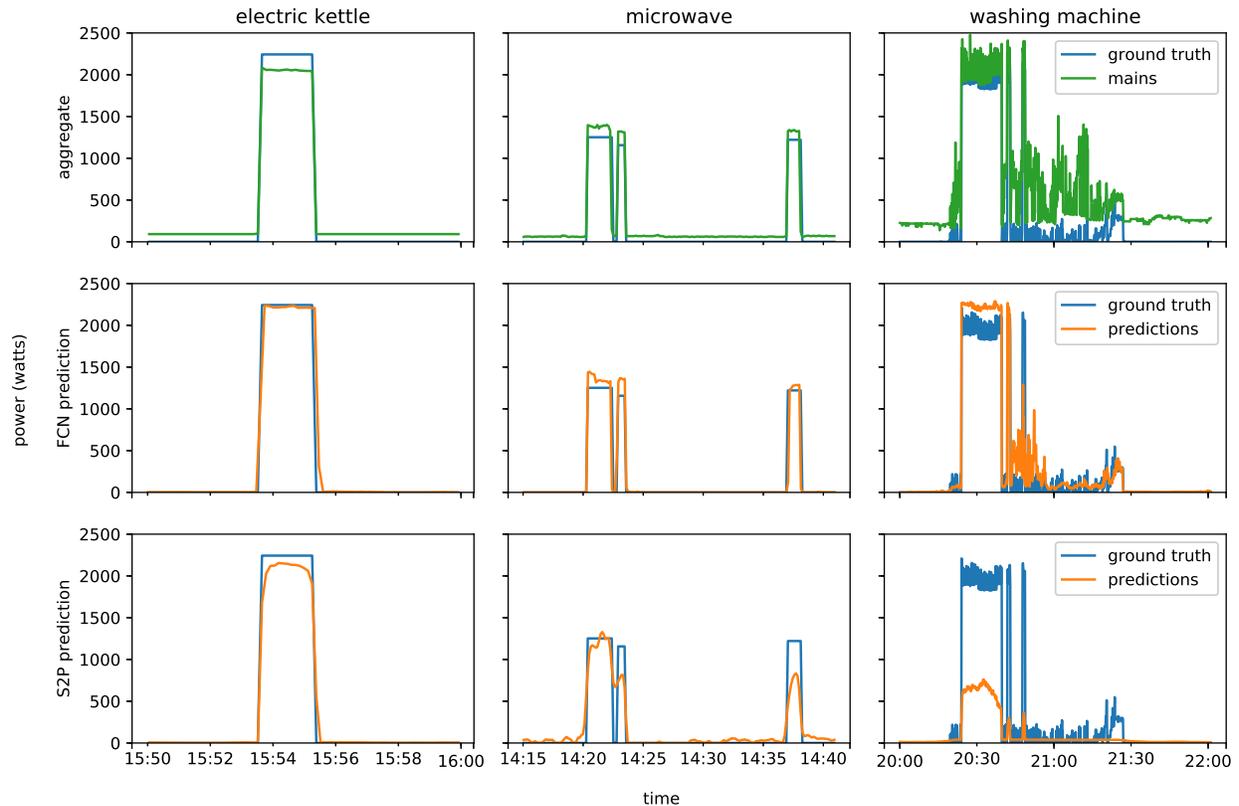}
\caption{Examples of predictions made by each model for the electric kettle, microwave, and washing machine.}
\end{figure}

The FCN is also much more efficient than the S2P model. As shown in table \ref{tab:efficiency-results}, training was  approximately 32 times faster and prediction was 43 times faster for the FCN. For a large scale system which would involve processing data from millions of homes, this speed up in prediction could significantly reduce the amount of hardware required. For some NILM applications, it may be desirable to perform disaggregation in the home using hardware with limited computational power which would require an efficient model.

\begin{table}[h]
\caption{Comparison of results from sequence to point network and fully convolutional network. The best result for each metric and appliance type is shown in bold.}
\label{tab:results}
\begin{tabular}{|l|rr|rr|rr|}

\hline
\textbf{}       & \multicolumn{2}{c|}{MAE (watt)} & \multicolumn{2}{c|}{SAE}      & \multicolumn{2}{l|}{\begin{tabular}[c]{@{}l@{}}Training time\\ (minutes)\end{tabular}} \\ \hline
Appliance       & FCN                 & S2P       & FCN           & S2P           & FCN                                            & S2P                                   \\ \hline
Electric Kettle & \textbf{10.4}       & 13.3      & \textbf{0.15} & 0.26          & \textbf{12.0}                                  & 541.8                                 \\
Microwave       & \textbf{4.4}        & 6.6       & \textbf{0.31} & 0.79          & \textbf{5.6}                                   & 261.0                                 \\
Washing Machine & \textbf{20.8}       & 28.8      & 0.39          & \textbf{0.34} & \textbf{34.9}                                  & 930.0                                 \\
Dishwasher      & \textbf{8.9}        & 19.4      & 0.11          & \textbf{0.05} & \textbf{5.6}                                   & 526.1                                 \\
Electric Shower & \textbf{10.0}       & 91.7      & \textbf{0.26} & 2.09          & \textbf{4.1}                                   & 134.6                                 \\
Electric Cooker & \textbf{34.8}       & 49.0      & 0.77          & \textbf{0.47} & \textbf{16.7}                                  & 138.6                                 \\ \hline
Mean            & \textbf{14.9}       & 34.8      & \textbf{0.33} & 0.67          & \textbf{13.2}                                  & 422.0                                 \\ \hline
\end{tabular}
\end{table}

\begin{table}[h]
\caption{Comparison of model efficiency for sequence to point network and fully convolutional network. Training and prediction were carried out using four Nvidia Titan X GPUs.}
\label{tab:efficiency-results}
\begin{tabular}{|l|l|l|}
\hline
                                 & FCN        & S2P                      \\ \hline
Parameter count                  & 428,801    & 30,708,249 \\
Mean training time (minutes)     & 13.2       & 422.0          \\
1 week prediction time (seconds) & 0.036      & 1.573      \\ \hline
\end{tabular}
\end{table}

\section{Conclusions and Future Work}

We have presented a fully convolutional neural network architecture for non-intrusive load monitoring. We have compared this network architecture to an existing state of the art model, and found that our proposed architecture achieved a lower error by two different metrics. Our model was also more computationally efficient than the existing model. The number of model parameters was reduced by a factor of 72, training time was reduced by a factor of 32, and prediction time was reduced by a factor of 43. 

The reduction in training time opens up many avenues for potential research, as the time taken for each experiment is reduced by an order of magnitude. For example, the model hyperparameters could be more easily optimised by random search when training time is reduced, or a large ensemble of models could be trained to achieve better performance. Other experiments which would involve training large numbers of models such as investigating the effect of changing the sampling rate or amount of training data could also be carried out more easily. 

There are several other paths of future research that may be worth investigating. Chen et al. \cite{chen18} proposed a new convolutional neural network architecture for NILM which makes use of gated linear units and residual connections. It would be interesting to compare this to our model and possibly investigate the effect of incorporating GLUs or residual connections into an FCN. Another interesting path could be to include additional information as inputs to the neural network, such as the types of appliance in each home. Fiterau et al. \cite{fiterau17} proposed a method of incorporating structured information into deep neural networks for time series modelling, which could be applied to our model.

\bibliographystyle{abbrv}
\bibliography{main}

\end{document}